\pdfoutput=1
\documentclass[11pt]{article}

\usepackage[]{acl}

\usepackage{times}
\usepackage{latexsym}
\usepackage{amsfonts,amssymb}
\usepackage[T1]{fontenc}
\usepackage[utf8]{inputenc}

\usepackage{microtype}

\usepackage{graphicx}
\usepackage{subcaption}
\usepackage{amsmath}
\title{Task-oriented Memory-efficient Pruning-Adapter}

\author{Guorun Wang \\
  Tongji University \\ 
  \normalsize{1950575@tongji.edu.cn} \\\And
  Jun Yang \\
  Tongji University \\
  \normalsize{junyang@tongji.edu.cn} \\\And
  Yaoru Sun \\
  Tongji University \\
  \normalsize{yaoru@tongji.edu.cn} }

\begin{document}
\maketitle
\begin{abstract}
The Outstanding performance and growing size of Large Language Models has led to increased attention in parameter efficient learning. The two predominant approaches are Adapters and Pruning. Adapters are to freeze the model and give it a new weight matrix on the side, which can significantly reduce the time and memory of training, but the cost is that the evaluation and testing will increase the time and memory consumption. Pruning is to cut off some weight and re-distribute the remaining weight, which sacrifices the complexity of training at the cost of extremely high memory and training time, making the cost of evaluation and testing relatively low.  So efficiency of training and inference can't be obtained in the same time. In this work, we propose a task-oriented Pruning-Adapter method that achieve a high memory efficiency of training and memory, and speeds up training time and ensures no significant decrease in accuracy in GLUE tasks, achieving training and inference efficiency at the same time.

\end{abstract}

\section{Introduction}

In recent years, there has been a significant increase in the size and complexity of Large Language Models, driven in part by the success of transformer-based models such as BERT\cite{devlin2018bert}, GPT-3\cite{brown2020language}, and T5\cite{2020t5}.  While these models have achieved SOTA (state-of-the-art) performance on a wide range of natural language processing tasks, such as GLUE\cite{wang2018glue}, their size has also become a major obstacle, particularly for deployment on smaller devices(GPU memory limited) such as mobile devices.

The sheer size of these models, often in the billions of parameters, requires significant amounts of memory and computational resource, which are often not available on such devices.  As a result, there has been growing interest in developing smaller, more efficient language models that can be deployed on resource-constrained devices without sacrificing too much in terms of performance.

In recent years, there has been a significant amount of research focused on improving the efficiency of LLM (Large Language Models). Two techniques that have gained considerable attention are Adapters and Pruning. Adapters\citep{hu2021lora,houlsby2019parameter,karimi2021compacter}  allow for the freezing of the model and the introduction of a new weight matrix on the side, which can result in a significant reduction in training time and memory usage. Similarly, Pruning\citep{voita-etal-2019-analyzing,Sajjad2020PoorMB,xia2022structured}  involves cutting off some weight and redistributing the remaining weight, which can lead to lower memory and training time costs, albeit at the expense of sacrificing the complexity of training. However, despite their effectiveness in improving the efficiency of LLM, both Adapters and Pruning suffer from limitations. Adapters, while reducing training time and memory usage, can increase the time and memory consumption of evaluation and testing. On the other hand, while Pruning can lower memory and inference time costs, it requires extremely high memory and training time to begin with. These limitations make them difficult to achieve both efficiency in training and inference simultaneously.

In this work, we propose a task-oriented Pruning-Adapter method and show that our method can achieve training and inference efficiency at the same time. \footnote{We will release the code soon.}
Our key insight is to jointly combine the advantages of Adapters and Pruning together both in training and testing. In summary, the contributions of this paper are fourfold as below:
\begin{itemize}
  \item We calculate the Task-oriented Importance, our studies have shown that the sensitivity of different Transformer structures to different types of data varies greatly.  
  \item We propose Importance-oriented Pruning, using Task-oriented Importance to prune Multihead Attention structures that are less important to the model.
  \item We propose Importance-oriented Rank-Varied LoRA, that we assign higher ranks to more important blocks, taking into account their importance in the  specific task.
  \item Our proposed method achieves a high memory efficiency of  training and memory, and speeds up training time and ensures no significant decrease in accuracy in GLUE tasks.
\end{itemize}

\section{Related work}
\subsection{Transformers}
Transformers\citep{NIPS2017_3f5ee243} has been widely used in NLP tasks, such as GLUE and open domain QA. It is composed of $L$(e.g., 12)  blocks and each consists a multi-head self-attention (MHA) layer, and a feed-forward (FFN) layer.

Each MHA layer which has $N_h$ heads(e.g., 12) and each FNN which consists of an up-projection $W_{up}$ and a down-projection layer $W_{down}$ takes input $X$ and outputs.

$$
MHA(X) = Concat (head_1,...,head_{N_h})W^O,
$$
$$
head_i = Attention(W^Q_i,W^K_i,W^V_i,X),
$$
$$
FNN(X)=relu(XW_{up})W_{down},
$$
where $W^Q_i$,$W^K_i$,$W^V_i \in \mathbb{R}^{d\times d_h}$ denote the 
query, key and value matrices respectively,  $W^O \in \mathbb{R}^{d\times d}$ denotes the output matrix, and $W_{up} \in \mathbb{R}^{d\times d_f}$ ,$W_{down} \in \mathbb{R}^{d_f\times d}$(normally $d_f$ = $4d$). 
Here $d$ denotes the model size(hidden size) (e.g., 768) and $d_h$ = $d/N_h$ denotes the output dimension of each head (e.g., 64). 
\subsection{Large Language pretrained Models and Tasks}
Nowadays, large pretrained models\citep{devlin2018bert,liu2019roberta,brown2020language,2020t5} with extraordinary performances on NLP tasks\citep{wang2018glue,wang2019superglue,izacard2020leveraging} emerge, however, these models that trained on enormous amounts of data have expensive time and GPU memory cost both in training and inference.

The intrinsic dimension\citep{aghajanyan2020intrinsic} is proposed as an objective function measures the minimum number of parameters needed to reach satisfactory solutions to the respective objective.
And normally simpler tasks and models that have perform better have relatively low intrinsic dimension, and intrinsic dimension decreases with training.

\subsection{Adapter}
Adapter\citep{hu2021lora,houlsby2019parameter,karimi2021compacter} is a way to freeze the model and give it a new, trainable weight matrix. The adapter method can significantly reduce the time and memory of training, but the cost is that the evaluation and testing will increase the time and memory consumption.

The adapter layer generally uses a down-projection project the input to a lower-dimensional space, followed by a nonlinear activation function $f(\cdot)$, and a up-projection to recover it. These adapters are added to the original matrix:
$$
output=\phi (X) + f(XW{}'_{down})W{}'_{up},
$$
where $\phi(\cdot)$ is the frozen transformer layer and $W{}'_{down}\in \mathbb{R}^{d\times r}$ and $W{}'_{up}\in \mathbb{R}^{r\times d}$ are restricted by the bottleneck dimension $r$. So adapters train very small parameter but there is no significant decline in performance. So far as to Adamix \citep{wang2022adamix}, which proposed Mixture of adapter, has performance better than full finetune. 

And from a unified view\citep{JunxianHe2021TowardsAU}, Prefix tuning\citep{li2021prefix} is derived as a special form of adapter, which reveals connection between prefix tuning and related work such as Prompt-tuning\citep{lester2021power}.

For instance, LoRA applies this equation to the query and value in the multi-head attention layer.\citep{hu2021lora} It adds a new pathway beside the original PLM, which simulates intrinsic rank by multiplying two matrices $A$ and $B$, where $A$ is responsible for dimension reduction and $B$ is responsible for dimension expansion, and the middle layer has a dimension of $r$. During downstream task training, the other parameters of the model are fixed, and only the weight parameters of the two new matrices are optimized.  The results of the PLM and the added pathway are added together to obtain the final result (the input and output dimensions of the two pathways are consistent), i.e.,
$$h=Wx+BAx$$. 
Usually, the weight parameters of the first matrix A are obtained by a Gaussian function, while the parameters of the second matrix B are set as zero matrix, so that the added pathway BA equals to zero at the beginning of training, and has no impact on the result, which is the same as the result of the original PLM.  During inference, the results of the two parts are added together as $h=Wx+BAx=(W+BA)x$, so as long as the trained matrix product $BA$ is added to the original weight matrix $W$ as new weight parameters to replace the original PLM's $W$, it will not significantly increase additional computing resources.

\subsection{Pruning}
Pruning\citep{voita-etal-2019-analyzing,Sajjad2020PoorMB,xia2022structured} is also emerging as it is to cut off some weight and re-distribute the remaining weight. Pruning has various methods such as Head pruning, Layer pruning etc., all of which apply variable mask $z$ to the equations of transformers. For instance,
$$
head_i = Z_{head}^{(i)} Attention(W^Q_i,W^K_i,W^V_i,X),
$$
It can be found that prune's approach sacrifices the complexity of training at the cost of extremely high memory and training time, making the cost of evaluation and testing relatively low.

These pruning method has different granularity, and in order to obtain higher acceleration ratio and lower accuracy loss, and alleviate the problem of high training cost, Coarse- and Fine-grained Pruning\citep{xia2022structured} was proposed to attach importance on pruning on  various granularity. 

Besides, due to the task specificity of most of the pruning method, some work explore the transfering ability cross task. Only 0.5\% of the pre-trained model parameters need to be modified per task.\citep{guo2020parameter}

\subsection{Parameter Importance}

It has been found that applying method on FNN randomly works poorly, because some columns(rows) contribute more than the others to model performance.\citep{zuo-etal-2022-moebert}

The importance score\citep{8953464} measures parameter importance. Intuitively, the more sensitive a model parameter is to the data, the more important it is\citep{zuo-etal-2022-moebert}, which can be reflected in the difference of the gradient in FNN layer.

Beyond that, this insight could be used in MHA layer.\citep{michel2019sixteen} Instead of backward the model, they use head\ mask $\xi $ of MHA to calculate the importance of MHA heads:

$$
I_h=\mathcal{E}_{x\sim X}  \left | \frac{\partial \mathcal {L}(x)}{\partial \xi}  \right | 
$$
$X$ is a small sample of training dataset, in this way, we could calculate the importance at a very low cost.

\begin{figure*}[ht]
  \centering
  \begin{subfigure}[b]{0.3\textwidth}
    \includegraphics[width=\textwidth]{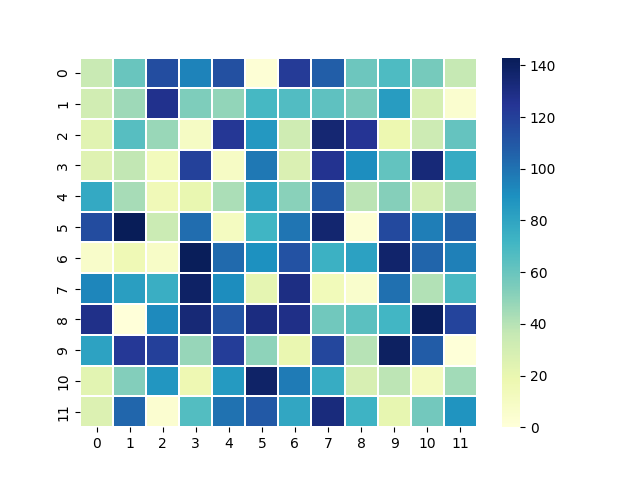}
    \caption{Mnli importance}
    \label{fig:1}
  \end{subfigure}
  \hfill
  \begin{subfigure}[b]{0.3\textwidth}
    \includegraphics[width=\textwidth]{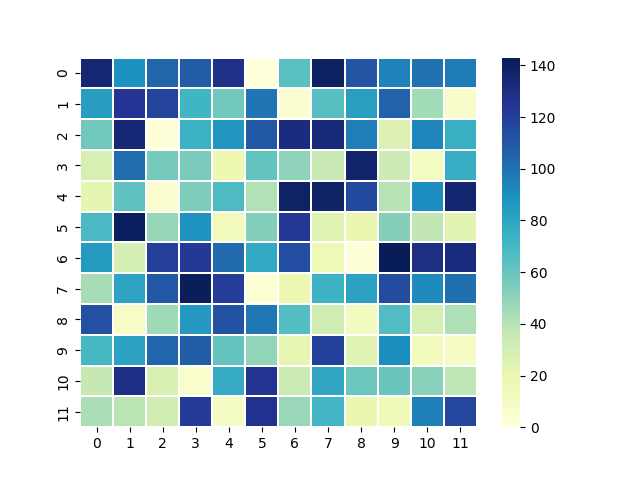}
    \caption{mrpc importance}
    \label{fig:2}
  \end{subfigure}
  \hfill
  \begin{subfigure}[b]{0.3\textwidth}
    \includegraphics[width=\textwidth]{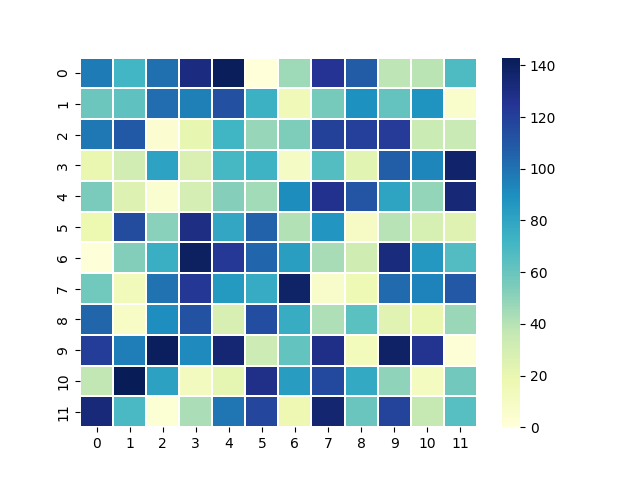}
    \caption{qnli importance}
    \label{fig:3}
  \end{subfigure}
  \hfill
  \begin{subfigure}[b]{0.3\textwidth}
    \includegraphics[width=\textwidth]{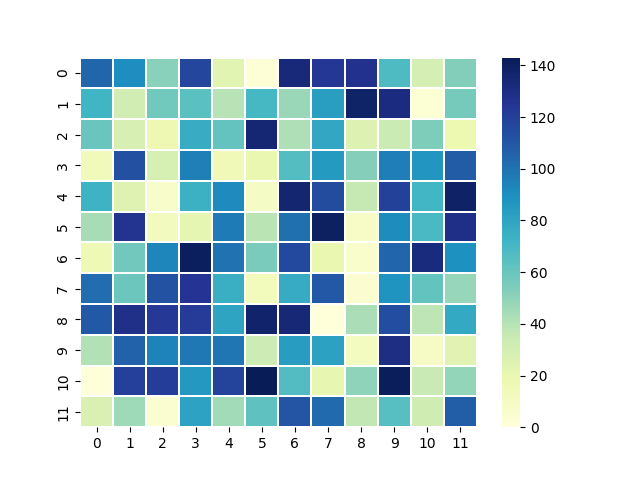}
    \caption{rte importance}
    \label{fig:4}
  \end{subfigure}
  \hfill
  \begin{subfigure}[b]{0.3\textwidth}
    \includegraphics[width=\textwidth]{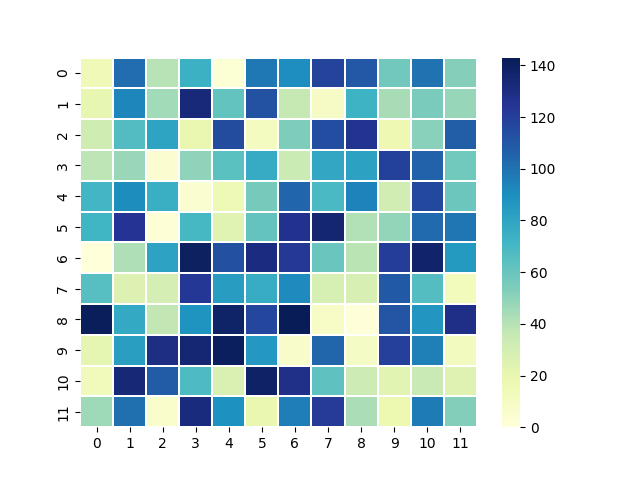}
    \caption{sst2 importance}
    \label{fig:5}
  \end{subfigure}
  \hfill
  \begin{subfigure}[b]{0.3\textwidth}
    \includegraphics[width=\textwidth]{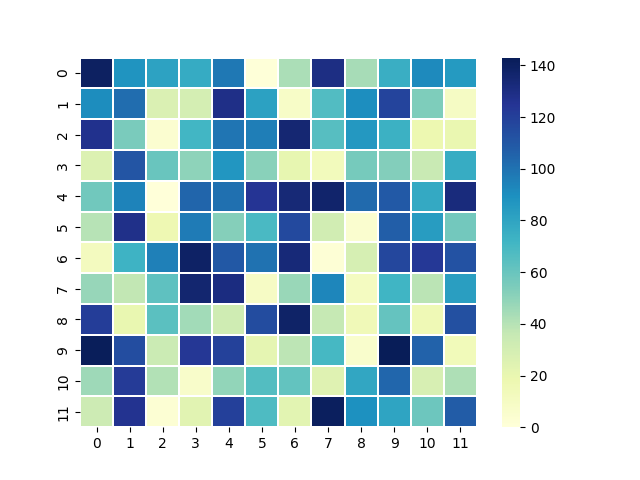}
    \caption{wnli importance}
    \label{fig:5}
  \end{subfigure}
  
  \caption{Importance of different data sets: we could see the transformer model is significantly more sensitive to different tasks. }
  \label{fig:Importance}
\end{figure*}

\section{Method}

\subsection{Task-oriented Importance}
The Transformer architecture\citep{NIPS2017_3f5ee243} has become a popular choice for a wide range of natural language processing tasks due to its ability to effectively model long-range dependencies.  

However, our studies have shown that the sensitivity of different Transformer structures to different types of data varies greatly.  
We used \cite{michel2019sixteen}'s method, 
they use head\ mask $\xi $ of MHA to calculate the importance of MHA heads:

% \begin{equation}
% \label{eq:quadratic}
% \begin{aligned}
%     I_h=\mathcal{E}_{x\sim X}  \left | \frac{\partial \mathcal {L}(x)}{\partial \xi}  \right | \\
% \end{aligned}
% \quad\text{and}\quad
% \begin{aligned}
%     I_H' = \frac{I_h}{\left\| \mathbf{I_h} \right\|_{2} + \epsilon}
% \end{aligned}
% \end{equation}
    % \begin{aligned}
    %     I_h=\mathcal{E}_{x\sim X}  \left | \frac{\partial \mathcal {L}(x)}{\partial \xi}  \right | 

    %     I_H' = \frac{I_h}{\left\| \mathbf{I_h} \right\|_{2} + \epsilon}
    % \end{aligned}

\begin{equation}
    \begin{split}
    I_h &=\mathcal{E}_{x\sim X}  \left | \frac{\partial \mathcal {L}(x)}{\partial \xi}  \right |, \quad \\
    I_H^{\prime} &= \frac{I_h}{\left\| \mathbf{I_h} \right\|_{2} + \epsilon}
    \end{split}
    \label{eq:quadratic}
\end{equation}

where $\left\| \mathbf{I_h} \right\|_{2} = \sqrt{\sum_{i=1}^{n} I_{h_i}^2}$ is the L\_2 norm of $I_h$, and $n$ is the length of the vector $I_h$. We introduce a small constant $\epsilon$ to make the operation legal. It should be noticed that in the calculation of expectation, we divide the importance by the total number of tokens in the data set $X$ to achieve a fairer measure.

We can normalize it further to obtain the final version of importance.

\begin{equation}\label{eq:quadratic}
    I_H = \frac{I_H' - min(I_H')}{max(I_H') - min(I_H')}
\end{equation}

% $\left\| \mathbf{I_h} \right\|_{2} = \sqrt{\sum_{i=1}^{n} I_h_i^2}$

and then we show the varied-importance in $12 \times 12$ map, shown in Figure \ref{fig:Importance}, where 12 denotes the number of blocks and the number of heads in each block.

This results in varying importance weights for the model's parameters across different tasks.  For example, a certain Transformer architecture may be more sensitive to syntax-related tasks, while another may perform better on tasks that require semantic understanding.  Therefore, it is crucial to carefully choose the appropriate Transformer structure for each task to achieve optimal performance.

It is worth noting that our operation to explore the importance of features only requires a small portion of data and does not require backward propagation through the entire model. Instead, we only need the gradient of the head mask, rather than the gradient of the entire model weights. This makes it particularly efficient in terms of lightweight model design. In fact the whole process can be be performed on the CPU, and the time and memory overhead of the computation is negligible.

\begin{figure*}[ht]
  \centering
  \includegraphics[width=0.5\textwidth]{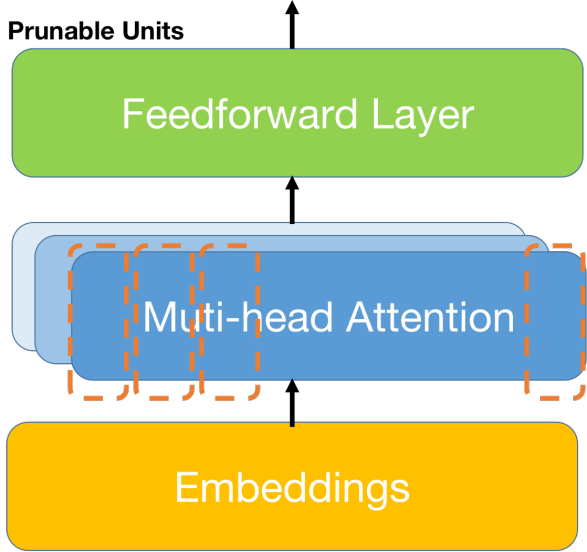}
  \caption{Prunable structure in our model: We do our pruning on the query, key, value and output of Multi-head Attention. In the figure, the orange dotted box is prunable units.}
  \label{fig:prune}
\end{figure*}
\subsection{Importance-oriented Pruning}
In the Transformer model, we calculate the weight of each head (144 in total) and determine which head to prune. The head mask tensor is used to prune off headers that are weighted below a certain threshold,  which may need to be adjusted for specific task requirements. In our experiments, we set it as 100, which means we prune 44 heads in our model. 

We have previously explored the formula for the attention of a single head:
\begin{equation}\label{eq:attention}
head_i = Attention(W^Q_i,W^K_i,W^V_i,X)
\end{equation}

During pruning, if the attention head of $i$ needs to be pruned, its weight matrix $W_i^Q$, $W_i^K$, $W_i^V$and the weight of output layer $W_{i, :}^O$ are all set to 0 in equation \ref{eq:attention} to achieve pruning operation. In practical implementation, the weight matrix of all the heads that need to be pruned and the weight of the output layer can be formed into a large sparse matrix, which is calculated only at the positions that need to be calculated.

Our implementation can be shown in Figure \ref{fig:prune}.

\begin{figure*}[ht]
  \centering
  \includegraphics[width=0.8\textwidth]{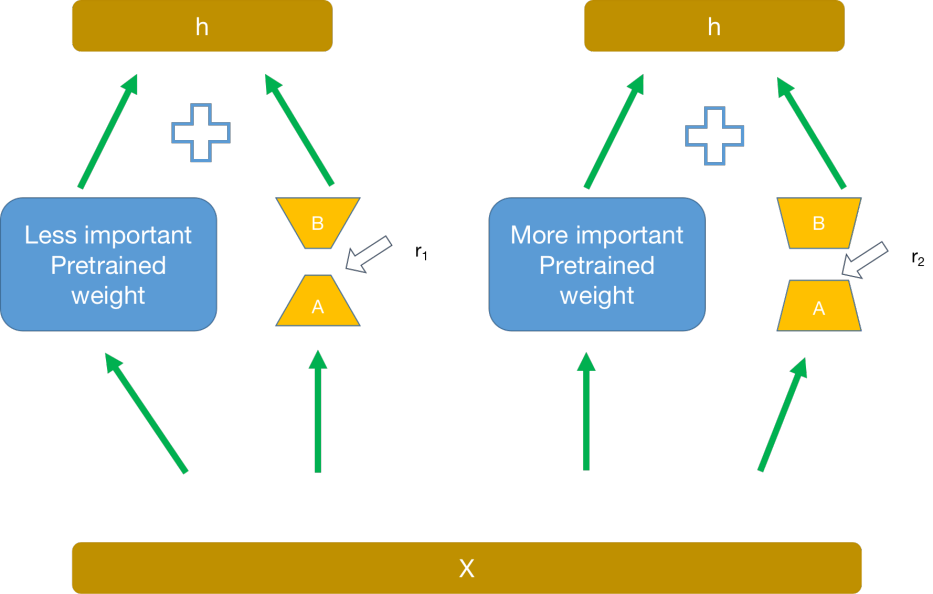}
  \caption{LoRA structure: We give lower rank matrixes to less important pretrained weight and give higher rank matrixes to more important pretrained weight}
  \label{fig:lora}
\end{figure*}

\subsection{Importance-oriented Rank-Varied LoRA}

As already described in the above, based on the Transformer model, we can measure the importance of each attention head as well as each block. Besides, the importance of each block can vary depending on the task. 

In LoRA equation\citep{hu2021lora}, we introduce two bypass weight:
\begin{equation}\label{eq:lora}
    h=Wx+BAx=(W+BA)x
\end{equation}

the choice of rank in the two bypass weights of LoRA  can also affect its performance. Empirically, a higher rank often leads to better performance as it allows for more trainable parameters. As such, we propose to assign higher ranks to more important blocks, taking into account their importance in the given task.

So the equation becomes:

\begin{equation}\label{eq:lora}
    h=W'x+B'A'x=(W'+B'A')x
\end{equation}
where $W'$ is pruned weight and $B'$ and $A'$ are Rank-Varied matries. The more important matrices $B'$ and $A'$ will have a higher rank. It should be noticed that the dimension of $W'$ changes, since $W'$ is pruned, which leads to different sizes of $B'$ and $A'$. Our structure can be shown in Figure \ref{fig:lora}.

\section{Experiments}
\subsection{Setup}
\subsubsection{Datasets and Baseline}
We evaluate our approach on six GLUE tasks\cite{wang2018glue}. GLUE tasks include mnli\cite{N18-1101}, mrpc\cite{neculoiu2016learning}, qnli, rte\cite{dagan2006pascal}, sst2, wnli.
We use bert-base-uncased model as our baseline. All of our code implementations are based on Huggingface\cite{wolf2019huggingface}.

\subsubsection{Training Setup}
In our experiments, we trained 30 epochs with the learning rate of $2 \times 10^{-5}$ (weight decay rate is 0.01), and the batch size is 32. We took the 4 most important blocks and give them a higher rank, which is 8, in each tasks. And the rest 8 less important blocks are given a lower rank, which is 4. We apply LoRA method in query, key, value and output layer in each block. The trainable param is layer norm layer and LoRA layer.

\subsubsection{Complexity Calculation}
In our experiments, the model parameters, memory cost and trainable parameters were calculated in train and inference. We also used Deepspeed's Flop Profiler \cite{rasley2020deepspeed}to aid in the calculations.

\subsection{Main result}
\subsubsection{Excellent accuracy performance}

We evaluate our model on GLUE task and the accuracy does not decline significantly, and in some ways equalled or even surpassed that of the no-pruning method. As shown in Table \ref{table:accuracy}

\begin{table*}
\centering
\begin{tabular}{lllllll}
\hline
 & \textbf{mnli} & \textbf{mrpc} & \textbf{qnli} & \textbf{rte} & \textbf{sst2} & \textbf{wnli}\\
\hline
Full finetune & 84.56\% & 80.14\% &91.54\% & 62.45\% & 92.43\% & 40.85\% \\
LoRA & 77.24\% & 68.38\%  & 88.05\% & 60.65\% & 91.63\% & 43.66\% \\
Prune-LoRA & \textbf{78.16\%} & 68.38\% &87.28\% &59.93\% &91.06\% & \textbf{56.34\%}  \\
\hline
\end{tabular}
\caption{\label{table:accuracy}
Excellent accuracy performance: The accuracy does not decline significantly, and in some ways equalled or even surpassed that of the no-pruning method.
}

\end{table*}

\subsubsection{Memory Efficiency}
Our method has excellent performance in memory cost in both training and inference. As an adapter model, we are actually more efficient than full finetune in inference memory consumption. In addition, there is an excellent decline in the inference model parameters, which is conducive to the inference that the Large Language Model is more efficient in small devices (GPU memory is limited), shown in Table \ref{table:Memory}. 

In addition, compared with the full finetune model, our training parameters are significantly reduced, only about 0.3\% of the original, which indicates that our model can also be trained on small devices. It is worth noting that, We give lower rank matrixes to less important pretrained weight and give higher rankmatrixes to more important  pretrained weight, so it has slightly more negligible (0.06\%) parameters than the traditional LoRA method, as shown in Table \ref{table:Param}.

\begin{table}
\centering
\begin{tabular}{lll}
\hline
 & \textbf{GPU memory} & \textbf{Params} \\
\hline
Full finetune & 418.7 MB & 109.48 M  \\
LoRA & 419.6 MB & 109.7 M    \\
Prune-LoRA & \textbf{392.4 MB} & \textbf{101.29 M }  \\
\hline
\end{tabular}
\caption{\label{table:Memory}
Memory performance of the inference : We calculated the memory cost of the model during inference and used the Deepspeed Flops Profiler to calculate the complexity of the model. Our method has excellent performance.
}

\end{table}

\begin{table*}
\centering
\begin{tabular}{llll}
\hline
 & \textbf{Model Param} & \textbf{Trainable Param} & \textbf{Proportion} \\
\hline
Full finetune & 109.48 M & 109.48 M & 100\% \\
LoRA & 109.7 M  & \textbf{259.6 K} &  \textbf{0.24\%} \\
Prune-LoRA & \textbf{101.29 M} & 308.7 K & 0.3\%  \\
\hline
\end{tabular}
\caption{\label{table:Param}
Trainable Param: Since our approach gives relatively high rank to more important blocks, our trainable parameters are a small increase over the conventional LoRA approach, but are still lightweight and efficient for the overall model.
}

\end{table*}

\subsubsection{Training Time Efficiency}
Since the total parameters and trainable parameters of our model are sharply reduced compared with those of full finetune, our training time becomes more efficient and we can train the model more quickly, as shown in Table \ref{table:Time}.

\begin{table}
\centering
\begin{tabular}{cc}
\hline
 & \textbf{Relative Training Time}  \\
\hline
Full finetune & 49   \\
LoRA & 36    \\
Prune-LoRA & \textbf{32}    \\
\hline
\end{tabular}
\caption{\label{table:Time}
The relative length ratio of training time, taking the mnli task as an example. 
}

\end{table}

In conclusion, our method is lightweight and efficient in the inference process such as training and training. As an adapter model, it combines some advantages of prune and inference, making the memory consumption and training time of training and training have the best performance.

\section{Future Work}
First of all, our model only carries out lightweight operation on the multihead attention, but not on the embedding and fnn layer. However, for transformer, FNN layer also takes a large proportion, so the next step is to apply our method to FNN layer. Second, although we have obtained the optimization of memory consumption and training time on train inference and inference, with respect to deepspeed inference time test, our method has certain deceleration according to the traditional method. We guess that huggingface transformer uses mask when it uses pruning. Although it reduces memory overhead, it cannot improve the forward efficiency of the model. More profound internal mechanism and mathematical formula derivation need to be further explored.

\section{Conclusion}

In this paper, combining the advantages of adapter and pruning, we proposed a task-based and memory-efficient Pruning-Adapter method that achieve a high memory efficiency of  training and memory, and speeds up training time and ensures no significant decrease in accuracy in tasks. We hope that future research continues this line of work, given that apply such methods to FNN layers will could possibly keep improving efficiency.

\bibliography{ref}

\begin{thebibliography}{28}
\expandafter\ifx\csname natexlab\endcsname\relax\def\natexlab#1{#1}\fi

\bibitem[{Aghajanyan et~al.(2020)Aghajanyan, Zettlemoyer, and
  Gupta}]{aghajanyan2020intrinsic}
Armen Aghajanyan, Luke Zettlemoyer, and Sonal Gupta. 2020.
\newblock Intrinsic dimensionality explains the effectiveness of language model
  fine-tuning.
\newblock \emph{arXiv preprint arXiv:2012.13255}.

\bibitem[{Brown et~al.(2020)Brown, Mann, Ryder, Subbiah, Kaplan, Dhariwal,
  Neelakantan, Shyam, Sastry, Askell et~al.}]{brown2020language}
Tom Brown, Benjamin Mann, Nick Ryder, Melanie Subbiah, Jared~D Kaplan, Prafulla
  Dhariwal, Arvind Neelakantan, Pranav Shyam, Girish Sastry, Amanda Askell,
  et~al. 2020.
\newblock Language models are few-shot learners.
\newblock \emph{Advances in neural information processing systems},
  33:1877--1901.

\bibitem[{Dagan et~al.(2006)Dagan, Glickman, and Magnini}]{dagan2006pascal}
Ido Dagan, Oren Glickman, and Bernardo Magnini. 2006.
\newblock The pascal recognising textual entailment challenge.
\newblock In \emph{Machine Learning Challenges. Evaluating Predictive
  Uncertainty, Visual Object Classification, and Recognising Tectual
  Entailment: First PASCAL Machine Learning Challenges Workshop, MLCW 2005,
  Southampton, UK, April 11-13, 2005, Revised Selected Papers}, pages 177--190.
  Springer.

\bibitem[{Devlin et~al.(2018)Devlin, Chang, Lee, and
  Toutanova}]{devlin2018bert}
Jacob Devlin, Ming-Wei Chang, Kenton Lee, and Kristina Toutanova. 2018.
\newblock Bert: Pre-training of deep bidirectional transformers for language
  understanding.
\newblock \emph{arXiv preprint arXiv:1810.04805}.

\bibitem[{Guo et~al.(2020)Guo, Rush, and Kim}]{guo2020parameter}
Demi Guo, Alexander~M Rush, and Yoon Kim. 2020.
\newblock Parameter-efficient transfer learning with diff pruning.
\newblock \emph{arXiv preprint arXiv:2012.07463}.

\bibitem[{He et~al.(2021)He, Zhou, Ma, Berg-Kirkpatrick, and
  Neubig}]{JunxianHe2021TowardsAU}
Junxian He, Chunting Zhou, Xuezhe Ma, Taylor Berg-Kirkpatrick, and Graham
  Neubig. 2021.
\newblock Towards a unified view of parameter-efficient transfer learning.
\newblock \emph{Learning}.

\bibitem[{Houlsby et~al.(2019)Houlsby, Giurgiu, Jastrzebski, Morrone,
  De~Laroussilhe, Gesmundo, Attariyan, and Gelly}]{houlsby2019parameter}
Neil Houlsby, Andrei Giurgiu, Stanislaw Jastrzebski, Bruna Morrone, Quentin
  De~Laroussilhe, Andrea Gesmundo, Mona Attariyan, and Sylvain Gelly. 2019.
\newblock Parameter-efficient transfer learning for nlp.
\newblock In \emph{International Conference on Machine Learning}, pages
  2790--2799. PMLR.

\bibitem[{Hu et~al.(2021)Hu, Shen, Wallis, Allen-Zhu, Li, Wang, Wang, and
  Chen}]{hu2021lora}
Edward~J Hu, Yelong Shen, Phillip Wallis, Zeyuan Allen-Zhu, Yuanzhi Li, Shean
  Wang, Lu~Wang, and Weizhu Chen. 2021.
\newblock Lora: Low-rank adaptation of large language models.
\newblock \emph{arXiv preprint arXiv:2106.09685}.

\bibitem[{Izacard and Grave(2020)}]{izacard2020leveraging}
Gautier Izacard and Edouard Grave. 2020.
\newblock Leveraging passage retrieval with generative models for open domain
  question answering.
\newblock \emph{arXiv preprint arXiv:2007.01282}.

\bibitem[{Karimi~Mahabadi et~al.(2021)Karimi~Mahabadi, Henderson, and
  Ruder}]{karimi2021compacter}
Rabeeh Karimi~Mahabadi, James Henderson, and Sebastian Ruder. 2021.
\newblock Compacter: Efficient low-rank hypercomplex adapter layers.
\newblock \emph{Advances in Neural Information Processing Systems},
  34:1022--1035.

\bibitem[{Lester et~al.(2021)Lester, Al-Rfou, and Constant}]{lester2021power}
Brian Lester, Rami Al-Rfou, and Noah Constant. 2021.
\newblock The power of scale for parameter-efficient prompt tuning.
\newblock \emph{arXiv preprint arXiv:2104.08691}.

\bibitem[{Li and Liang(2021)}]{li2021prefix}
Xiang~Lisa Li and Percy Liang. 2021.
\newblock Prefix-tuning: Optimizing continuous prompts for generation.
\newblock \emph{arXiv preprint arXiv:2101.00190}.

\bibitem[{Liu et~al.(2019)Liu, Ott, Goyal, Du, Joshi, Chen, Levy, Lewis,
  Zettlemoyer, and Stoyanov}]{liu2019roberta}
Yinhan Liu, Myle Ott, Naman Goyal, Jingfei Du, Mandar Joshi, Danqi Chen, Omer
  Levy, Mike Lewis, Luke Zettlemoyer, and Veselin Stoyanov. 2019.
\newblock Roberta: A robustly optimized bert pretraining approach.
\newblock \emph{arXiv preprint arXiv:1907.11692}.

\bibitem[{Michel et~al.(2019)Michel, Levy, and Neubig}]{michel2019sixteen}
Paul Michel, Omer Levy, and Graham Neubig. 2019.
\newblock Are sixteen heads really better than one?
\newblock \emph{Advances in neural information processing systems}, 32.

\bibitem[{Molchanov et~al.(2019)Molchanov, Mallya, Tyree, Frosio, and
  Kautz}]{8953464}
Pavlo Molchanov, Arun Mallya, Stephen Tyree, Iuri Frosio, and Jan Kautz. 2019.
\newblock \href {https://doi.org/10.1109/CVPR.2019.01152} {Importance
  estimation for neural network pruning}.
\newblock In \emph{2019 IEEE/CVF Conference on Computer Vision and Pattern
  Recognition (CVPR)}, pages 11256--11264.

\bibitem[{Neculoiu et~al.(2016)Neculoiu, Versteegh, and
  Rotaru}]{neculoiu2016learning}
Paul Neculoiu, Maarten Versteegh, and Mihai Rotaru. 2016.
\newblock Learning text similarity with siamese recurrent networks.
\newblock In \emph{Proceedings of the 1st Workshop on Representation Learning
  for NLP}, pages 148--157.

\bibitem[{Raffel et~al.(2020)Raffel, Shazeer, Roberts, Lee, Narang, Matena,
  Zhou, Li, and Liu}]{2020t5}
Colin Raffel, Noam Shazeer, Adam Roberts, Katherine Lee, Sharan Narang, Michael
  Matena, Yanqi Zhou, Wei Li, and Peter~J. Liu. 2020.
\newblock \href {http://jmlr.org/papers/v21/20-074.html} {Exploring the limits
  of transfer learning with a unified text-to-text transformer}.
\newblock \emph{Journal of Machine Learning Research}, 21(140):1--67.

\bibitem[{Rasley et~al.(2020)Rasley, Rajbhandari, Ruwase, and
  He}]{rasley2020deepspeed}
Jeff Rasley, Samyam Rajbhandari, Olatunji Ruwase, and Yuxiong He. 2020.
\newblock Deepspeed: System optimizations enable training deep learning models
  with over 100 billion parameters.
\newblock In \emph{Proceedings of the 26th ACM SIGKDD International Conference
  on Knowledge Discovery \& Data Mining}, pages 3505--3506.

\bibitem[{Sajjad et~al.(2020)Sajjad, Dalvi, Durrani, and
  Nakov}]{Sajjad2020PoorMB}
Hassan Sajjad, Fahim Dalvi, Nadir Durrani, and Preslav Nakov. 2020.
\newblock Poor man's bert: Smaller and faster transformer models.
\newblock \emph{ArXiv}, abs/2004.03844.

\bibitem[{Vaswani et~al.(2017)Vaswani, Shazeer, Parmar, Uszkoreit, Jones,
  Gomez, Kaiser, and Polosukhin}]{NIPS2017_3f5ee243}
Ashish Vaswani, Noam Shazeer, Niki Parmar, Jakob Uszkoreit, Llion Jones,
  Aidan~N Gomez, \L~ukasz Kaiser, and Illia Polosukhin. 2017.
\newblock \href
  {https://proceedings.neurips.cc/paper/2017/file/3f5ee243547dee91fbd053c1c4a845aa-Paper.pdf}
  {Attention is all you need}.
\newblock In \emph{Advances in Neural Information Processing Systems},
  volume~30. Curran Associates, Inc.

\bibitem[{Voita et~al.(2019)Voita, Talbot, Moiseev, Sennrich, and
  Titov}]{voita-etal-2019-analyzing}
Elena Voita, David Talbot, Fedor Moiseev, Rico Sennrich, and Ivan Titov. 2019.
\newblock \href {https://doi.org/10.18653/v1/P19-1580} {Analyzing multi-head
  self-attention: Specialized heads do the heavy lifting, the rest can be
  pruned}.
\newblock In \emph{Proceedings of the 57th Annual Meeting of the Association
  for Computational Linguistics}, pages 5797--5808, Florence, Italy.
  Association for Computational Linguistics.

\bibitem[{Wang et~al.(2019)Wang, Pruksachatkun, Nangia, Singh, Michael, Hill,
  Levy, and Bowman}]{wang2019superglue}
Alex Wang, Yada Pruksachatkun, Nikita Nangia, Amanpreet Singh, Julian Michael,
  Felix Hill, Omer Levy, and Samuel Bowman. 2019.
\newblock Superglue: A stickier benchmark for general-purpose language
  understanding systems.
\newblock \emph{Advances in neural information processing systems}, 32.

\bibitem[{Wang et~al.(2018)Wang, Singh, Michael, Hill, Levy, and
  Bowman}]{wang2018glue}
Alex Wang, Amanpreet Singh, Julian Michael, Felix Hill, Omer Levy, and Samuel~R
  Bowman. 2018.
\newblock Glue: A multi-task benchmark and analysis platform for natural
  language understanding.
\newblock \emph{arXiv preprint arXiv:1804.07461}.

\bibitem[{Wang et~al.(2022)Wang, Agarwal, Mukherjee, Liu, Gao, Awadallah, and
  Gao}]{wang2022adamix}
Yaqing Wang, Sahaj Agarwal, Subhabrata Mukherjee, Xiaodong Liu, Jing Gao,
  Ahmed~Hassan Awadallah, and Jianfeng Gao. 2022.
\newblock Adamix: Mixture-of-adapter for parameter-efficient tuning of large
  language models.
\newblock \emph{arXiv preprint arXiv:2205.12410}.

\bibitem[{Williams et~al.(2018)Williams, Nangia, and Bowman}]{N18-1101}
Adina Williams, Nikita Nangia, and Samuel Bowman. 2018.
\newblock \href {http://aclweb.org/anthology/N18-1101} {A broad-coverage
  challenge corpus for sentence understanding through inference}.
\newblock In \emph{Proceedings of the 2018 Conference of the North American
  Chapter of the Association for Computational Linguistics: Human Language
  Technologies, Volume 1 (Long Papers)}, pages 1112--1122. Association for
  Computational Linguistics.

\bibitem[{Wolf et~al.(2019)Wolf, Debut, Sanh, Chaumond, Delangue, Moi, Cistac,
  Rault, Louf, Funtowicz et~al.}]{wolf2019huggingface}
Thomas Wolf, Lysandre Debut, Victor Sanh, Julien Chaumond, Clement Delangue,
  Anthony Moi, Pierric Cistac, Tim Rault, R{\'e}mi Louf, Morgan Funtowicz,
  et~al. 2019.
\newblock Huggingface's transformers: State-of-the-art natural language
  processing.
\newblock \emph{arXiv preprint arXiv:1910.03771}.

\bibitem[{Xia et~al.(2022)Xia, Zhong, and Chen}]{xia2022structured}
Mengzhou Xia, Zexuan Zhong, and Danqi Chen. 2022.
\newblock Structured pruning learns compact and accurate models.
\newblock \emph{arXiv preprint arXiv:2204.00408}.

\bibitem[{Zuo et~al.(2022)Zuo, Zhang, Liang, He, Zhao, and
  Chen}]{zuo-etal-2022-moebert}
Simiao Zuo, Qingru Zhang, Chen Liang, Pengcheng He, Tuo Zhao, and Weizhu Chen.
  2022.
\newblock \href {https://doi.org/10.18653/v1/2022.naacl-main.116} {{M}o{EBERT}:
  from {BERT} to mixture-of-experts via importance-guided adaptation}.
\newblock In \emph{Proceedings of the 2022 Conference of the North American
  Chapter of the Association for Computational Linguistics: Human Language
  Technologies}, pages 1610--1623, Seattle, United States. Association for
  Computational Linguistics.

\end{thebibliography}
\bibliographystyle{acl_natbib}

\appendix

\end{document}